# Protecting Society from AI Misuse: When are Restrictions on Capabilities Warranted?

**Markus Anderljung**[1], **Julian Hazell**[1,2]

[1] Centre for the Governance of AI, Oxford, United Kingdom

[2] Oxford Internet Institute, University of Oxford, Oxford, United Kingdom

markus.anderljung@governance.ai, julian.hazell@governance.ai

## ABSTRACT

Artificial intelligence (AI) systems will increasingly be used to cause harm as they grow more capable. In fact, AI systems are already starting to be used to automate fraudulent activities, violate human rights, create harmful fake images, and identify dangerous toxins. To prevent some misuses of AI, we argue that targeted interventions on certain capabilities will be warranted. These restrictions may include controlling who can access certain types of AI models, what they can be used for, whether outputs are filtered or can be traced back to their user, and the resources needed to develop them. We also contend that some restrictions on non-AI capabilities needed to cause harm will be required. Though capability restrictions risk reducing use more than misuse (facing an unfavorable Misuse-Use Tradeoff), we argue that interventions on capabilities are warranted when other interventions are insufficient, the potential harm from misuse is high, and there are targeted ways to intervene on capabilities. We provide a taxonomy of interventions that can reduce AI misuse, focusing on the specific steps required for a misuse to cause harm (the Misuse Chain), and a framework to determine if an intervention is warranted. We apply this reasoning to three examples: predicting novel toxins, creating harmful images, and automating spear phishing campaigns.

## 1 Introduction

Recent advances in AI technologies have been accompanied by concerns about how these systems could be misused to cause harm [10]. Though these harms were once speculative, they are now becoming increasingly felt. Drug discovery algorithms can be used to detect novel toxins [65]. Large language models have started being used to design malware[1] and automate fraud [16]. Visual recognition systems are being used to identify and persecute minority populations [47]. Image generation models are being used to create pornographic deepfakes without subjects' consent [72]. Automated armed drones have been used on the battlefield [30,31] and may soon be used by non-state actors or for human rights violations. As these harms increase, so too will calls to address them.

The growing list of AI misuses has motivated debate around what interventions (if any) are warranted for preventing misuse of AI systems. For example, in September 2022, U.S. Congresswoman Anna Eshoo called for an investigation into cases of misuse of Stable Diffusion, an image-generation model released by Stability AI just one month prior. Eshoo argued in her letter that the model has the capability to create "real world harms" such as political propaganda, violent imagery, child pornography, copyright violations, and disinformation, and should therefore be "governed appropriately" [19]. As Eshoo further noted, AI models are often dual-use, with the potential for both harm and benefit [20].

Decision-makers across AI developers, legislative bodies, regulatory agencies, and social media platforms must therefore navigate a precarious balancing act when attempting to govern powerful AI systems – one that effectively prevents misuse without interfering too much

---

[1] Oftentimes, the malware produced by such systems is not very sophisticated, but language models do significantly lower the barrier to creating it [36, 15].

with beneficial uses, resulting in a positive *Misuse-Use Tradeoff*.

Here, we seek to help steady this balancing act. We begin by surveying how AI can be misused. We further provide decision-makers with a framework for thinking about what AI interventions are possible and which may be warranted. Building on the framework, we claim there are cases that warrant interventions that modify what AI capabilities exist, who has access to them, and what kind of access is granted. We conclude by exploring three case studies – AI for toxin generation, harmful image production, and spear phishing.

For the purposes of this report, we define "misuse" as "the intentional use of AI to achieve harmful outcomes" [10]. This definition excludes accidents and incompetent use of AI, which lack the intentionality of cases of misuse. What counts as "harmful consequences" is value-laden. "Misuse" is therefore an undeniably contentious and political concept. As such, we aim to stick with reasonably clear cases of misuse that are likely to be seen as harmful to large swathes of society. An "intervention," in turn, describes an action or policy that has the goal of addressing misuse. Interventions work by making sure misuse does not happen in the first place (or affects less people), is less harmful if it does happen, or is appropriately responded to after the fact.

## 2 How AI Can Be Misused

Large language models (LLMs) could be used to increase the speed and scale of text-based cyber attacks such as spear phishing. Frontier language models have the ability to write large amounts of sophisticated text for as little as cents.[2] This combination of scale and sophistication could supplant human operators for large-scale spear phishing campaigns. Some experts have proposed using AI-powered cyber defense systems to help reduce these kinds of risks [10], while others have expressed the need for oversight of such defensive systems [60].

Moreover, LLMs may enable malicious actors to generate increasingly sophisticated and persuasive propaganda and other forms of misinformation [28,33]. Similar to automated phishing attacks, LLMs could increase both the scale and sophistication of mass propaganda campaigns. The use of large language models to automate propaganda can result in a higher number of propagandists as the reliance on manual labor is decreased, thus reducing the overall costs of these campaigns. Language models can also change actors' behavior by introducing novel tactics, such as real-time content generation, and improve existing tactics such as cross-platform testing. Finally, the content of propaganda campaigns may also change as messages can be made more credible if models are fine-tuned to mimic effective propagandists [28]. Proposed solutions aimed at mitigating these harms include ensuring labs build models to be truthful, encouraging governments to impose controls on AI hardware and data collection, and fostering collaboration between AI developers and content platforms to create tools and processes aimed at detecting AI-generated content [28].

Authoritarian governments could misuse AI to improve the efficacy of repressive domestic surveillance campaigns. The Chinese government has increasingly turned to AI to improve its intelligence operations, including facial and voice recognition models and predictive policing algorithms [46]. Notably, these technologies have been used for the persecution of the Uyghur population in the Xinjiang region. This persecution might constitute crimes against humanity, according to a recent UN report [64]. In response, it has been suggested that democratic countries coordinate to design export controls that stifle the spread of these technologies to authoritarian regimes [46].

AI could be used to create lethal autonomous weapon systems (LAWS) with significant misuse potential. Some critics have argued that LAWS could enable human commanders to commit criminal acts without legal accountability [34], be used by non-state actors to commit acts of terrorism [25], and violate human rights [51]. As the technology has matured, LAWS have been increasingly adopted by various militaries. For example, autonomous weapon systems might have already been used on the battlefield in Ukraine and Libya [37]. Yet even though some nation states have released declarations on the responsible use of AI for military operations [13], international negotiations aiming to constrain the adoption of LAWS have largely stalled [62]. In response, calls for non-proliferation of LAWS have been suggested as a complement to norms governing their use [61].

Finally, advanced image generation models also have a range of potential misuses. These models could be used to create harmful content, including depictions of nudity, hate, or violence [42]. Moreover, they could be utilized to reinforce biases and subject individuals or groups to indignity. There is also the potential for these models to be used for exploitation and harassment, such as by removing articles of clothing from pre-existing images or memorizing an individual's likeness without their consent. Furthermore, image generation models could be used to spread disinformation by depicting political figures in unfavorable contexts. To avoid these harms, various proposals have been suggested (and indeed, implemented), such as removing explicit content from training data, filtering texts prompts that violate terms of use, implementing use rate limits to prevent at-scale abuse, adding visual image signatures to detect AI-generated content, and using monitoring and human review to detect policy violations [42].

---

[2] OpenAI's API allows for the generation of 750 words at a cost of $0.002 with their GPT 3.5 Turbo model.

# 3 Types of Intervention to Address Misuse

Interventions to address misuse can be categorized by looking at the process by which a misuse of AI causes harm: (i) Some actor needs to carry out the misuse, for which they need to have the relevant AI and non-AI capabilities. Accordingly, interventions can modify capabilities to reduce misuse. (ii) Once the misuse has been carried out, it causes harm via some route, such as exposing individuals to some harmful content. Interventions can seek to mitigate harm. (iii) After misuse has taken place, interventions can respond to the harm. We will call these steps the "*Misuse Chain*."

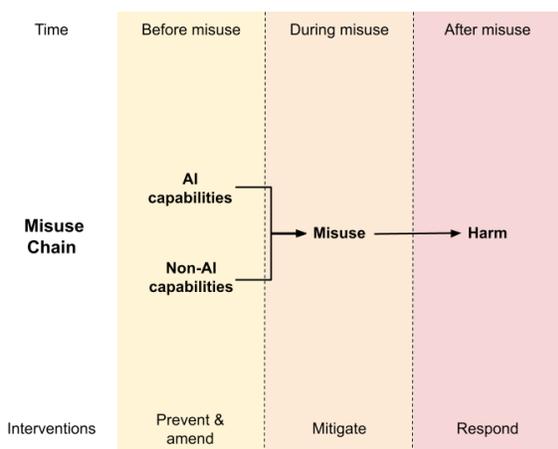

**Figure 1: An Illustration of the Misuse Chain.**

## 3.1 Modify Capabilities

To carry out the misuse, an actor first needs to acquire the necessary capabilities. Often, the actor will combine some AI capability with other non-AI capabilities to carry out the misuse. For example, producing AI-designed toxins requires the outputs of an AI system as well as non-AI inputs, such as the physical ingredients needed to synthesize the toxins. Below, we will focus on possible interventions geared at modifying what capabilities (AI and non-AI) exist, and which actors have access to them.

*3.1.1 Interfering with Misuse via Capabilities.* Interventions at the capabilities stage can focus on AI models, the resources required to develop and run them, or any non-AI inputs needed to cause harm. In turn, these capabilities-focused interventions can reduce misuse by impacting the scope and scale of potential harm, in addition to improving the efficacy of future interventions at later stages of the Misuse Chain.

*3.1.1.1 Interventions Aimed at AI Models.* Interventions aimed at AI models can impact what models are produced, who gets access to them, and what they are allowed to be used for. Adjusting what kinds of models are developed, developers can train models that perform poorly at misuse-relevant tasks. In doing so, the model is made less dual-use: it becomes more useful for positive versus negative uses [53]. For example, an image generation model can be made worse at producing sexual images of someone's likeness without their consent if the dataset used to train the model is scrubbed of sexual images. OpenAI's DALLE-2 had violent and sexual images filtered from its training dataset, for example, and Stable Diffusion 2.0 had not safe for work-content removed from its dataset [42,59]. Similarly, LLMs could be trained to be honest, helpful, and harmless, rather than merely capable of producing human-like text [7].

It is also possible to intervene on who has access to the model and how they can use it. This can be done via *usage restrictions*, making it difficult or impossible to use an AI system for specific purposes. For example, content filters can be introduced for generative models (e.g. large language models and text-to-image models). Such filters can ensure the model does not provide certain kinds of outputs. They can focus on the inputs (e.g. checking if the prompt includes certain sensitive keywords) or the outputs (e.g. having a classifier assess whether the output includes restricted categories). Today, these filters are fairly imprecise: as of the writing of this paper, OpenAI's DALL-E 2 responds to the prompt "Joe Biden announcing his campaign, wearing a funny hat" with "It looks like this request may not follow our content policy." In the future, as the quality of classifiers increases, these filters might become more precise and sophisticated.

*User restrictions* can also be introduced, where attempts are made to ensure specific actors – e.g. those believed most likely to misuse the technology – have limited access to it. For example, AI developers could put in place Know Your Customer processes and check their users against a list of known scammers, potential terrorists, or those on the US Entity List, for example. Users who appear to be misusing their model or otherwise violating their terms of service could lose access.

All of these interventions may require that the model is accessed via a structured access scheme, where the user can interact with the model through an API, but does not have full access to its weights [55]. Though many of the above measures could be implemented in models that users have full access to, such access would likely allow users to disable and circumvent the measures. Sometimes circumventing these measures requires trivial effort: in early versions of Stable Diffusion, the safety filter could be disabled by removing a few lines of code.

Further, it is important that the potential impact and capabilities of AI systems are assessed and that such assessments inform decisions about deployment and, increasingly, development. This includes asking the question: what dangerous capabilities does this model have and what can we do to ensure those capabilities are not misused? Such information is crucial for understanding the scale of misuse and designing appropriate interventions in turn. AI developers are in a privileged position to make such assessments before deployment and provide relevant information to society, including as part of published papers [44], and are starting to recognize this fact [4,36,52]. Further, they can invite external red teams to scrutinize their models and provide assessments before model release or deployment [11]. However, we should not expect these assessments to be fully comprehensive, as the uses to which AI systems can be put – including misuses – can be difficult to foresee, especially for large self-supervised models [26]. As such, society needs to continually monitor and evaluate the capabilities and impacts of AI models.

*3.1.1.2 Interventions Aimed at Inputs to AI Capabilities.* Interventions can also focus on the resources that are required to develop and deploy AI models: compute, algorithmic insights, data, and talent.

Developing and deploying AI models often requires specialized computing infrastructure. Where such infrastructure is difficult to get a hold of – for example, if the model is large enough that it cannot easily be run or trained on readily available computing hardware – there may be room for intervention. These interventions could focus on who has access. For example, many compute providers, including cloud providers as well as sellers of AI-relevant hardware, already likely implement various Know Your Customer processes to ensure compliance with US Entity List requirements. These processes could be expanded to include other actors who are likely to misuse compute. Further, the US government introduced wide-ranging export controls on AI-relevant chips going to China in October 2022 to thwart what it considers Chinese misuses of AI capabilities, such as using AI for domestic surveillance and human rights abuses [14].

Access to computing resources can also be amended depending on what it is being used for. For example, there have been increasing calls for cloud compute providers to conduct human rights assessments, investigating the risk that their compute provision aids human rights abuses [39]. The world's three largest cloud providers, Amazon, Microsoft, and Google, have indicated that their business practices are informed by UN Guiding Principles on Business and Human Rights [3,57,69]. Such principles encourage businesses to carry out human rights due diligence when making decisions that have the potential to cause adverse human rights impacts – in this example, providing cloud compute for end-users or end-use cases that could violate human rights. Further, as the impact of AI systems increases, assurances that compute is not aiding misuse may need to scale with the amount of compute provided [12].

Interventions aimed at reducing access to certain algorithmic insights are likely to be blunt, but may nonetheless be warranted at times. Certain knowledge may cause more harm than good if released widely [56]. AI developers and researchers could choose to not publish certain discoveries should it seem that doing so would cause sufficient harm. AI research publication venues could implement ethics reviews and require that researchers reflect on the potential harmful impacts, including misuse, of their research [5,32,49]. In extreme cases, should there be insights in AI development that could cause severe harm if widely distributed, governments might consider introducing secrecy orders on relevant patents [24].

*3.1.1.3 Interventions Aimed at AI-Complementing Capabilities.* Some misuses of AI are only feasible when paired with other non-AI capabilities. For example, the ability to design novel toxins is only problematic if such toxins can then be produced and distributed. Much of the export control regime connected to the Chemical Weapons Convention focuses on reducing access to prohibited chemicals and precursors thereof, rather than limiting access to information on their toxicity and how they might be produced.

In the case of AI-generated influence campaigns on social media platforms, misuse will require access to accounts that appear authentic. If AI-generated content cannot be widely disseminated, large-scale influence campaigns utilizing such content would be rendered ineffective. Access to this necessary non-AI capability could be undermined by requiring accounts be authenticated via IDs [28] or by requiring accounts pay a small subscription fee [2].

*3.1.2 Amending Misuse.* Interventions on capabilities are not only valuable inasmuch as they prevent actors from misusing AI systems, or reduce the harm caused by the misuse when it does occur. Interventions in the capabilities stage of the Misuse Chain can also *amend* the misuse, teeing up other interventions down the line.

As an example of amending misuse, the provider could try ensuring text generated by their system can be traced back to its source, should law enforcement present a warrant.

Doing so could enable more effective responses to misuse that would discourage further misuse attempts.

There are a number of ways in which a provider of an LLM could help others identify whether a piece of text was AI generated or produced by a particular system. One option is to attempt to introduce watermarks into the outputs of the system [58]. One method of doing so involves making the model more statistically likely to use certain phrases or words, in a way that is unnoticeable to humans, but can be picked up by a detector provided a long enough sequence of text. One weakness of this approach is that it might be possible to circumvent by having another system paraphrase the original text [1]. Another option is to keep a database of outputs from the model that can then be matched to text on the internet. This approach also has some limitations, such as raising privacy concerns, being computationally intensive, and sometimes producing false positives, but may nonetheless be helpful [1].

## 3.2  Mitigate Harm

Once the misuse has been carried out, it causes harm via some route, such as by exposing individuals to some content.

To mitigate harm, interventions can focus on identifying and stopping the misuse's spread or reduce the harm of exposure. Taking AI-enabled influence operations on social media as an example, interventions could focus on reducing the number of people who are exposed to AI-generated disinformation. This could be done by identifying AI-generated content and reducing its virality, more generally reducing the virality of political or divisive content, or making it harder to automate the posting of such content. Interventions could also focus on the harm that results from such distribution, for example by tagging certain content as AI-generated, introducing fact-checking measures, introducing friction to sharing articles without reading them first, or increasing users' media literacy [28].

Various forms of fingerprinting or hash matching [15,29] are now used for identifying, removing, or reducing the spread of prohibited content (for example, copyrighted material or known child sexual abuse images) on social media platforms. Machine learning systems can also be trained on large datasets to classify new or previously unseen prohibited material [9,40]. Social platforms can leverage this power to take a more interventionist role by downranking certain content. Facebook, for instance, tweaked its algorithm in 2018 to demote content that is deemed close to violating its community standards [74].

Interventions could also focus on the harm that results from such distribution, for example by tagging certain content as AI-generated – as proposed in the forthcoming EU AI Act [21]. Malicious actors could also be deprived of information they need to increase the effectiveness of their misuse. For example, Twitter recently introduced a monthly fee for API access, claiming this was partly to cut off access to bots [70].

Often, offensive capabilities can also be used for defense. In the cybersecurity domain, for example, AI systems could be used to identify and exploit software vulnerabilities. But they can also be employed by defenders to detect and patch these vulnerabilities. With sufficient effort – assuming it is possible to construct vulnerability-free code – the defender could eventually become invulnerable to attacks [27]. Similarly, language models could be used to automate spear phishing attacks, but could also be used to identify and screen out attempts at such attacks.

Interventions at this stage could also amend the harm. For example, they could focus on ensuring information is collected that might aid responses to it. This might include measuring the extent of the harm from these misuses, which can inform policy. It might also include collecting information needed to identify and sanction the perpetrator of the harm.

## 3.3  Respond to Harm and Misuse

After some misuse has taken place, interventions can respond to it.

Interventions can seek to sanction the perpetrator of the harm, thereby disincentivizing misuse. For example, several jurisdictions have, over the past couple of years, introduced laws and voluntary frameworks for governing the production and distribution of non-consensual deepfake pornography [22,23]. Unauthorized access to computer networks can come with criminal charges in the US, with penalties including up to ten years in prison [68]. Sanctions often target attempts at the misuse action, regardless of whether it in fact caused harm; an incompetent assassin can still be charged with attempted murder.

Actors who did not engage in misuse but nonetheless enabled it could still be incentivized to invest in mitigation measures. For instance, in the US, the Fair Credit Billing Act (FCBA) offers consumers safeguards against credit card fraud by limiting their responsibility in the event of fraud or billing errors. If a credit card is used without authorization, the cardholder is allowed to challenge any charges exceeding $50 [67]. This rule incentivises financial institutions to invest in security measures [41]. In other cases, actors compensate harmed parties even without legal obligation. For example, e-commerce platforms might not be legally required to refund customers who have been

misled or scammed, yet still elect to offer purchase assurance to ensure customers remain satisfied.

Decisionmakers can also adjust policy and rules upon learning about the harm. For example, in 2019 a Harvard student used GPT-2 to submit 1,001 responses to an Idaho request for comments on its Medicaid program. The comments were taken seriously until the student informed the authorities [71]. Partly as a response, the US government's official portal for US federal public comments now includes security measures such as CAPTCHAs, as suggested by a bipartisan report documenting abuse of the US government's online commentary system [45].

# 4 When is an Intervention Warranted?

It is difficult to judge whether a particular intervention or set thereof is warranted. Attempts to reduce the misuse of AI systems almost always come at some cost: interventions may require the investment of significant funds and can impinge on freedoms or pose privacy concerns. Further, nearly all attempts to stop bad or unacceptable uses of AI also hinder positive uses; there is a *Misuse-Use Tradeoff*.

Hindering positive uses is no trivial matter. Doing so can present significant costs to society. In addition to the use itself being valuable, reducing such use can come with significant negative externalities. Decreased access to frontier models in AI for academics for fear they may be misused, for example, could significantly reduce society's ability to scrutinize and understand the limitations and potential impacts of increasingly powerful AI systems.

On the other hand, reducing misuse can also come with significant positive externalities. There exist analogous cases where preventing misuse brought wide benefits to society. Without spam filters, email would be less widely used and valuable. One study by researchers at Microsoft and Google estimated that internet users would encounter 300 times as many spam emails if firms did not invest in anti-spam technology [50].

In light of the above, we recommend that decisions be informed by estimates of interventions' Misuse-Use Tradeoff. To assess the Misuse-Use Tradeoff, we suggest that decision-makers consider the following ratios.

*4.1 Value Ratio.* First, consider the ratio of the disvalue of the misuses to the value of the uses. How do the harms from the misuses, including their negative externalities, compare to the benefits of the uses, including their positive externalities? This ratio is 1 if the disvalue of the misuses exactly matches the value of the uses. If the ratio is more than 1, the misuses are more harmful than the uses are beneficial, and vice versa if the ratio is below 1. The worse the misuses are compared to the value of uses, the more interventions are warranted.

The value of use is in turn a function of the number of uses and their average value. Similarly, the disvalue of the misuse is a function of the number of misuses and the average disvalue or harm that comes from those misuses. As such, the Value Ratio is equal to the ratio between the number of misuses and the number of uses, multiplied by the average disvalue of each misuse and value of each use.

*4.2 Targetedness Ratio.* Second, we must consider the ratio of how much the intervention affects misuse versus use. The more the intervention impacts the misuses without disturbing the uses, the better the case for it. The Targetedness Ratio can be defined as the percentage decrease in use value divided by the percentage decrease in the disvalue or harm from misuse.

We can further decompose the Targetedness Ratio into two parts. First, consider the True Positive-False Positive Ratio: what is the ratio between the chance that the intervention correctly identifies a misuse and the chance it mistakenly tags a use as a misuse? The better the intervention is at picking out the misuses and not picking up any of the uses, the stronger the case for it.

Second, we can look at the Effectiveness ratio. That is, how much does the intervention reduce the harm caused by the misuse compared to how much it reduces the value of any uses it affects? The more the intervention reduces the harm of the misuses it affects and the less it reduces the value of uses that get caught in the crossfire, the stronger the case for the intervention.

We can summarize the above with the following equation:

$$\mathbf{Misuse - Use\ Tradeoff}$$

$$= \text{Value Ratio} \cdot \text{Targetedness Ratio}$$

where

$$\text{Value Ratio} = \left[\frac{\#\text{ misuses}}{\#\text{ uses}} * \frac{\text{average misuse value}}{\text{average use value}}\right], \text{ and}$$

$$\text{Targetedness Ratio} = \left[\frac{\%\text{ true positives}}{\%\text{ false positives}} * \frac{\text{average effect on misuses}}{\text{average effect on uses}}\right]$$

# 5 Capability interventions can be untargeted

Though we believe capability restrictions will be necessary and warranted to address certain misuses, one might generally prefer interventions aimed at the harm or response stages of the Misuse Chain. Capability restrictions are often a blunt tool, naturally so because they are more causally distant from the downstream misuse. Interventions that minimize the harm of misuses or respond to them after the fact can be sensitive to more facts about the situation, such as the outputs that the actor has produced with a model or how those outputs have been used. This means that later interventions in the Misuse Chain are more likely to have a better Targetedness Ratio.

Partly for these reasons, society tends to deal with actors intentionally causing harm by focusing on the Mitigate and Respond parts of the Misuse Chain. Law enforcement tends to focus on finding and punishing crimes rather than preempting them. In the case of AI, the challenges of ensuring that LLMs are helpful, harmless, and honest provides a useful illustration [6]. Whether an output is harmless or not depends largely on context, context which the creator of the LLM often lacks. Answers to the question "what are the most effective ways to hack this network?" could be used nefariously or could be used by cybersecurity professionals to identify potential system vulnerabilities.

However, there are many exceptions to this pattern. Society has put in place many interventions on actors' access to capabilities in attempts to reduce misuse. Internet service providers block or throttle traffic to certain websites, such as ones used for piracy or other illegal activity. Commercially available drones come with preset "geofences", which describe virtual boundaries that, when crossed by the drone, trigger warnings and cause the drone to hover in place [18]. The development, possession, and use of chemical and biological weapons is governed by international treaties, such as the Chemical Weapons Convention and the Biological Weapons Convention. Further, many precursors to chemical weapons are also restricted. Certain chemicals used in the final stage of chemical weapons production are considered themselves to be chemical weapons under the convention, and are therefore regulated similarly to the final products [43].

# 6 When Interventions Aimed at Restricting Capabilities are Warranted

This section discusses qualitative factors that make it more likely that capabilities interventions are warranted. That is: where interventions at other stages are not sufficiently effective, where the harm from misuse is large, and where there are targeted interventions.

## 6.1 Where Interventions at Other Stages Are Not Sufficiently Effective

Interventions aimed at restricting capabilities are worth stronger consideration where interventions at other stages are less effective. Take nuclear weapons as an example. It is difficult to defend against a nuclear weapon. Compared with conventional weapons, nuclear weapons have the capacity to cause widespread destruction in comparatively miniscule time scales, and missile defense systems have the potential to enhance the likelihood of confrontation [48]. Similarly, states would prefer to not maintain stability via the deterrent effect of a retaliatory strike, given its accompanying risk of escalation. As such, we reduce access to the capability where possible via non-proliferation efforts.

This partly explains the US government's increasingly harsh export controls on AI chips and chip manufacturing tools going to China. Believing that such exports would be used for what the US government considers misuses and seeing that they can likely not intervene at later parts in the Misuse Chain, they intervene on Chinese access to AI-relevant compute. Notably, this same logic may come to be applied to other AI capabilities, such as trained AI models or certain datasets.

## 6.2 When the Harms from Misuse are Sufficiently Large

If the harms from the misuses of a model outweigh the benefits from its use, capability restrictions become far more appealing. The fact that such restrictions may be more likely to bring the value of the relevant system to zero is a blessing, not a curse. There are certain capabilities that we simply should prefer not to exist. AI systems specifically designed to create explicit deepfake content of any person's likeness, such as DeepNude, an AI application that uses neural networks to remove people's clothing in images, provide a compelling example. Similarly, AI models specifically designed to circumvent attempts to detect and stop misuse – say models designed to remove watermarks from AI-generated images – are likely to do more harm than good.

Further, if the harm from misuse are sufficiently large in absolute terms, interventions at the capabilities stage may also be warranted. Such harms should increase willingness to pay higher fixed costs to design an intervention with high targetedness. For instance, while AI algorithms for drug discovery could yield beneficial advances, they could also potentially be misused to design novel toxins. Even if the legitimate benefits of drug discovery outweigh such misuses, these malicious applications would be severe and deserve significant effort to thwart.

## 6.3 When an Intervention has Minimal Effects on Uses

Capability interventions that have minimal or no effects on uses – leading to a high Targetedness Ratio – are more desirable. Many important interventions at the capabilities stage aim not to directly stop the misuse, but to modify it in ways that boost the effectiveness of interventions later in the Misuse Chain. For example, interventions can aim to ensure that it is possible for other actors to detect whether an output is AI-generated or describe crucial features of the output. In the example of deep fakes, it may be necessary to determine if the content is explicit, of someone's likeness, and AI generated. They can also aim to ensure that the output's provenance is known. These features then enable interventions further down the Misuse Chain, such as tagging AI-generated content as such to better inform its viewers, and being able to sanction actors misusing AI capabilities. The key benefit behind these interventions is that they tend to have minimal or no effect on the use of the system: they are highly targeted.

Interventions at the capabilities stage can also be made more targeted by carefully employing structured access approaches [55]. For example, many large language models available via APIs apply filters to their outputs. While these filters risk being either over-inclusive or under-inclusive, they could be made more targeted by measuring user behavior across multiple outputs. Users who consistently produce content that looks inappropriate are more likely to be misusing the system and could be flagged for further investigation or have their access reduced.

Importantly, interventions of this kind are often most effective when their details (and sometimes even their existence) are not widely shared. Avoiding detection is easier if you know what the detection procedure is. The effectiveness of speed cameras is greatly reduced if drivers know where they are located. Similarly, legal systems will often include intentional ambiguity and room for judgment to allow courts, law enforcement, and regulators to enforce the spirit rather than the letter of the law. If the rules and means of detecting a breach are made overly precise, actors can make sure to precisely skirt the line, avoiding detection while still being able to carry out misuse. This is analogous to how tax authorities around the world tend not to give precise details about their methods of detecting tax fraud.

## 7 Case Studies

Below, we discuss three case studies of AI misuse where targeted capabilities restrictions are warranted: AI systems used to predict toxins, image generation models being used to create harmful content, and LLMs being used for spear phishing campaigns.

## 7.1 Toxin Prediction

In 2021, two researchers created a list of novel toxins using MegaSyn, a machine learning based de novo molecule generator used for drug discovery [66]. Normally, MegaSyn penalizes expected toxicity and rewards expected bioactivity. But the researchers wondered what would happen if they flipped the filter on MegaSyn – literally by swapping a '1' for a '0' and a '0' for a '1' – to produce toxic molecules. After running the modified system on a 2015 MacBook for a few hours, a list of over 40,000 toxins was produced, some of which were predicted to be more lethal than publicly known chemical warfare agents such as VX. Concerningly, the pair of researchers created MegaSyn with publicly available data and software [65].

While certain forms of capabilities restrictions might not entirely prevent actors from misusing these models, interventions at the capabilities stage might nonetheless be desirable. With chemical weapons, minimizing harm seems difficult. The attack surface is vast and harm can be severe. Responding to misuse might be possible, yet such sanctions would ideally occur at the earliest possible stages of misuse, such as when an actor is planning their attack. Given the high stakes, the ineffectiveness of interventions at later stages of the Misuse Chain, and the potential to reduce the dual-use nature of these sorts of AI systems, various capabilities restrictions appear warranted.

For example, structured access schemes could make these models less dual-use. Making it so that drug discovery models can only be accessed via an API would enable filters that prohibit outputs of compounds above some threshold of toxicity. Legitimate uses of such outputs – e.g. by researchers seeking to develop countermeasures to novel toxins – could be enabled by an approval process. Additionally, structured access schemes could enable the model's owners to keep a log of its outputs, and who created them. This could allow law enforcement to preempt attacks as well as track down perpetrators after harm occurs. Even if rogue actors could circumvent these restrictions by training their own models, reducing the number of actors capable of causing harm is still desirable.

Other non-AI interventions at the capabilities stage likely play an even more important role. For example, requirements could be made for companies providing chemical synthesis services to screen orders and cooperate with law enforcement in the event that a malicious actor attempts to order highly dangerous toxins or precursors thereof. Governance structures could also target the technologies and materials used to synthesize the resulting chemical agents to ensure they cannot be misused. There are existing efforts to prevent the production of large quantities of toxins, such as the Chemical Weapons

Convention. These efforts, and others seeking to prevent the widespread dispersal of harmful toxins, might need to be updated and strengthened as the number and severity of novel toxins increases.

## 7.2 Harmful Image Generation

Certain images, such as pornographic images of someone's likeness produced without their consent and fake images intended to mislead public opinion, can cause harm. While it has long been possible to use tools like Photoshop to manipulate images, there was less need for intervention in the past. The production of highly realistic fake images was time-consuming and the resulting images would typically reach a limited audience. However, AI systems capable of producing photorealistic images significantly lower the barriers to producing such images. Paired with the increased reach that images can have when disseminated on the internet and social media platforms, such content can cause significantly more harm than before.

One way of preventing harm from AI-generated images is to prevent malicious actors from acquiring the capabilities needed to generate such images in the first place. However, some interventions aimed at ensuring generative models do not produce harmful images are fraught with difficulties. Take for example the strategy of introducing content filters on natural language prompts used for image generation models such as DALL-E 2. Getting these filters right is difficult: such filters are likely to be both underinclusive (e.g. images with violent content can be generated by indirect prompting, such as "a horse lying on its side in a puddle of red liquid") and overinclusive (e.g. perhaps disallowing any prompt that includes the phrase "breast stroke"). As a result, these kinds of post-hoc filters could hinder legitimate use while still being vulnerable to exploitation by malicious actors. Most likely, they will produce many more false positives than true positives.

Nevertheless, these challenges are not sufficient to rule out all interventions at the capabilities stage. Interventions at the capabilities stage can still amend misuse, thereby improving the efficacy of other interventions at later stages of the Misuse Chain. For example, owners of an image-generation model that is queried via an API could insert invisible watermarks into the model's outputs. These watermarks could help social media platforms detect AI-generated content more reliably. Fingerprints could also be installed directly within image generation models. For example, researchers have demonstrated a scalable technique that applies a unique fingerprint to each image produced by a particular copy of a model's weights, so misuse can be attributed back to specific users [73]. With improved detection capabilities, platforms could mitigate harm by labeling AI-generated content as such or remove media that violates their terms of service. The proposed EU AI Act includes related provisions, requiring posters of AI generated content to disclose it as such [21]. Some platforms, like Facebook [8] and Twitter [63], already have policies to remove certain manipulated media, but improved AI detection could strengthen their ability to enforce these policies.

Finally, model creators could train highly capable classifier models that are designed to attribute content to the model that generated it, even if these models are open-source. By keeping these classifier models hidden behind APIs, malicious actors could face difficulties attempting to reliably evade detection. To prevent misuse, actors should adhere to a norm of only releasing generative models broadly once sufficient safeguards are in place, such as capable detection systems.

## 7.3 Spear Phishing

Using AI systems for phishing campaigns could pose significant threats to nation states, organizations, and individuals. In particular, LLMs could enhance hackers' ability to "spear phish", a tactic in which customized communication is sent from an ostensibly trustworthy source in order to trick recipients into revealing sensitive information. The analogy of a sniper is often invoked when describing spear phishing, as it is a highly targeted and precise method of attack, whereas ordinary phishing is seen as a broad and indiscriminate attack, akin to a shotgun blast. Jeh Johnson, former U.S. Secretary of the Department of Homeland Security, was quoted as saying that "the most devastating, intrusive attacks by the most sophisticated actors often originate with a simple act of spear phishing" [17].

Spear phishing is traditionally a time-consuming and labor-intensive process that can involve several steps, such as identifying high-value targets, conducting personalized research to gather relevant information on the target, and crafting a tailored message that appears to come from a trusted acquaintance. However, with the integration of AI, this process can be made more efficient. Even relatively simple AI systems can improve attackers' efficiency. For example, in 2018 researchers created an automated spear phishing system called SNAP_R that used a long short-term memory neural network to send phishing tweets tailored to targets' characteristics [54]. Though the tweets were typically short and unsophisticated, SNAP_R could send them significantly faster than a human operator, and with a similar click-through rate, according to a small experiment the authors conducted. Compared to the models used to create SNAP_R, large transformer-based language models are significantly more capable of generating human-sounding text. There are clear indications of large language models being used for spear phishing, as evidenced by discussion on dark web forums for cybercriminals [35].

Despite the potential LLMs afford for improving spear phishing, intervening at the capabilities stage could pose challenges. First, it would be difficult to tell whether a given piece of text is intended to be used for spear phishing, or a similar sounding, non-malicious form of communication, such as a marketing email. Preventing a language model from outputting such text would be technically difficult, and could noticeably interfere with positive use. Secondly, malicious actors with enough motivation would likely be able to fine-tune the model to output such messages nonetheless.

However, some intervention at the capabilities stage may still be warranted. New threats will require modifications to existing cybersecurity systems, and perhaps the creation of entirely new forms of defense. Crucially, certain capabilities interventions can make LLMs less capable of being used for spear phishing, and can boost interventions at later stages of the Misuse Chain. For example, one solution could be to ensure that advanced language models are only queryable via a structured access scheme, such as an API. Similar to interventions aimed at preventing harmful images from being generated, suspicious requests could be flagged and logged in a database. Users who consistently produce text that could be used for spear phishing could then be investigated or subject to usage restrictions by the model's owners.

Being able to identify AI generated content as such would be helpful for mitigating harm. A recent classifier produced by OpenAI showed a 26% true positive rate and a 9% false positive rate for AI generated [38], which may not be sufficiently targeted. However, research is exploring methods for watermarking AI generated text by making the model biased in favor of certain word choices or combinations thereof that a classifier, but not a human, would be able to detect [1]. Though this method might be circumventable via having another AI model paraphrase the output, it may stop less sophisticated actors. Further, such paraphrasing attempts could also potentially be thwarted by sharing the classifier with other LLM providers, allowing them to check whether users are using the system to remove watermarks.

Moreover, the most effective interventions for preventing spear phishing may focus on mitigating harm. Harm from AI-generated spear phishing could be mitigated by improving existing systems aimed at stopping spear phishing. For example, systems can use LLMs to scan and then flag suspicious messages, taking into account various metadata such as whether the sender is using an email similar to someone already in the target's contact list. These systems do not need to necessarily tell if an incoming message is AI-generated. On the contrary, identifying and protecting against harm, regardless of whether or not it is AI generated, could be a more robust defensive strategy over the long run. More research is needed to determine whether spear phishing attacks or defenses against them will gain more from advances in these technologies.

# 8   Conclusion

AI systems are already being misused across various domains, and as they become more capable and are deployed more broadly, the potential for misuse will grow. Decision-makers will feel compelled to intervene on such misuses, but choosing the right suite of interventions can be difficult. Interventions inevitably face the Misuse-Use Tradeoff. Despite this tradeoff, we argue that interventions aimed at the capabilities stage of the Misuse Chain will be increasingly warranted as the potential harms of AI misuse increase, as AI misuse becomes difficult to defend against in the other stages of the Misuse Chain, and as new techniques are created that can increase the targetedness of capability interventions.

To better prepare society for managing AI misuse, we encourage future research on a number of questions, including:

- **Determining the potential harm of high-risk misuses and what interventions will be warranted.** What misuses of AI are high-risk? In high-risk domains, how much harm could be caused by AI misuse? Given this, what interventions are warranted?
- **Generating empirical estimates of Misuse-Use Tradeoffs**. How can we estimate the Misuse-Use Tradeoff for interventions such as filters used for image-generation models and LLMs?
- **Developing techniques to help defend against misuse**. What systems can most effectively prevent misuse while favorably navigating the Misuse-Use Tradeoff?

## ACKNOWLEDGEMENTS

We are grateful for the valuable insights and feedback provided by Ben Garfinkel, Ella Guest, Anne le Roux, Alexis Carlier, David McCaffary, Anton Korinek, Elisabeth Seger, Sam Clarke, Loenie Koessler, Eoghan Stafford, Noemi Dreksler, Robert Trager, Emma Bluemke, Allan Dafoe, Jonas Schuett, Jonas Sandbrink, Gregory Lewis, Sean Ekins, Jess Whittlestone, and to participants in the Centre for the Governance of AI work-in-progress sessions. Particular thanks go to Ben Bucknall and Toby Shevlane.

We would also like to acknowledge ChatGPT and Claude for their useful suggestions.